\documentclass[10pt,twocolumn,letterpaper]{article}

\usepackage{iccv}
\usepackage{times}
\usepackage{epsfig}
\usepackage{graphicx}
\usepackage{amsmath}
\usepackage{amssymb}

\usepackage{multirow}
\usepackage{mathrsfs}
\usepackage{enumitem}
\usepackage{makecell}
\usepackage{tabulary}
\usepackage{pifont}
\usepackage{subfigure}

\usepackage[breaklinks=true,bookmarks=false]{hyperref}

\iccvfinalcopy % *** Uncomment this line for the final submission

\def\iccvPaperID{****} % *** Enter the ICCV Paper ID here

\ificcvfinal\fi

\begin{document}

\title{Multi-Source Domain Adaptation and Semi-Supervised Domain Adaptation with Focus on Visual Domain Adaptation Challenge 2019}

\author{Yingwei Pan, Yehao Li, Qi Cai, Yang Chen, and Ting Yao\\
         JD AI Reseach, Beijing, China\\
         {\tt\small \{panyw.ustc, tingyao.ustc\}@gmail.com}
}

\maketitle
\ificcvfinal\thispagestyle{empty}\fi

%%%%%%%%% ABSTRACT

\begin{abstract}
This notebook paper presents an overview and comparative analysis of our systems designed for the following two tasks in Visual Domain Adaptation Challenge (VisDA-2019): multi-source domain adaptation and semi-supervised domain adaptation.

\textbf{Multi-Source Domain Adaptation}: We investigate both pixel-level and feature-level adaptation for multi-source domain adaptation task, i.e., directly hallucinating labeled target sample via CycleGAN and learning domain-invariant feature representations through self-learning. Moreover, the mechanism of fusing features from different backbones is further studied to facilitate the learning of domain-invariant classifiers.
Source code and pre-trained models are available at \url{https://github.com/Panda-Peter/visda2019-multisource}.

\textbf{Semi-Supervised Domain Adaptation}: For this task, we adopt a standard self-learning framework to construct a classifier based on the labeled source and target data, and generate the pseudo labels for unlabeled target data. These target data with pseudo labels are then exploited to re-training the classifier in a following iteration. Furthermore, a prototype-based classification module is additionally utilized to strengthen the predictions.
Source code and pre-trained models are available at \url{https://github.com/Panda-Peter/visda2019-semisupervised}.

\end{abstract}

\section{Introduction}
Generalizing a model learnt from a source domain to target domain, is a challenging task in computer vision field. The difficulty originates from the domain gap \cite{yao2012predicting} that may adversely affect the performance especially when the source and target data distributions are very different.  An appealing way to address this challenge would be unsupervised domain adaptation (UDA) \cite{cai2019exploring,saito2018maximum,zhang2018fully}, which aims to utilize labeled examples in source domain and the large number of unlabeled examples in the target domain to generalize a target model. Compared to UDA which commonly recycles knowledge from single source domain, a more difficult but practical task (i.e., multi-source domain adaptation) is proposed in \cite{peng2018moment} to transfer knowledge from multiple source domains to one unlabeled target domain. In this work, we aim at exploiting both pixel-level and feature-level domain adaptation techniques to tackle this challenge problem. In addition, another task of semi-supervised domain adaptation \cite{daume2010frustratingly,yao2015semi} is explored here when very few labeled data available in the target domain.

\begin{figure}[!tb]
\vspace{-0.0in}
	\centering {\includegraphics[width=0.49\textwidth]{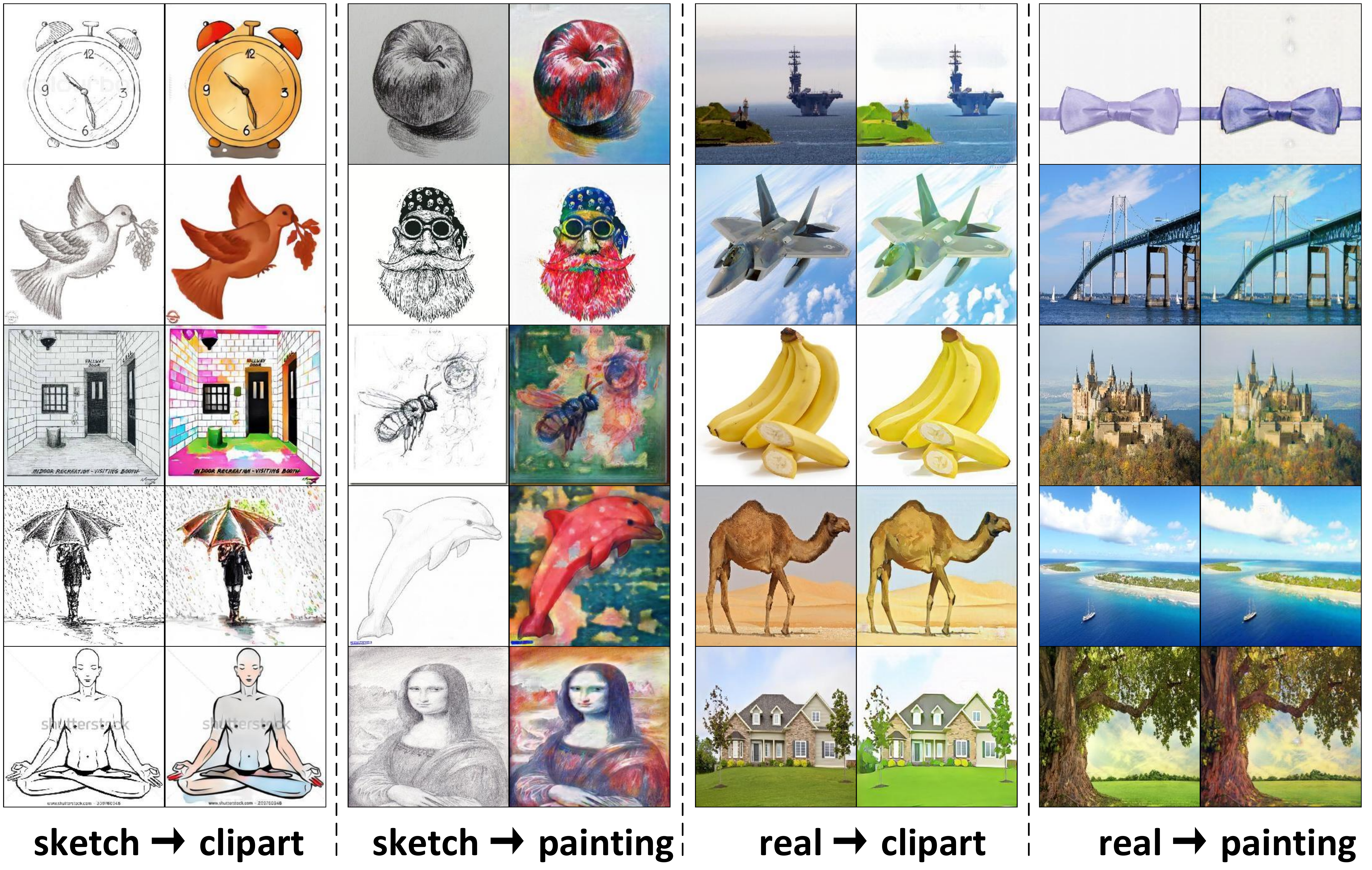}}
	\caption{\small Examples of Pixel-level adaptation between source domains (sketch and real) and target domain (clipart/painting) via CycleGAN in multi-source domain adaptation task.}
	\label{fig:0}
	\vspace{-0.2in}
\end{figure}

\begin{figure*}[!tb]
\vspace{-0.10in}
	\centering {\includegraphics[width=0.86\textwidth]{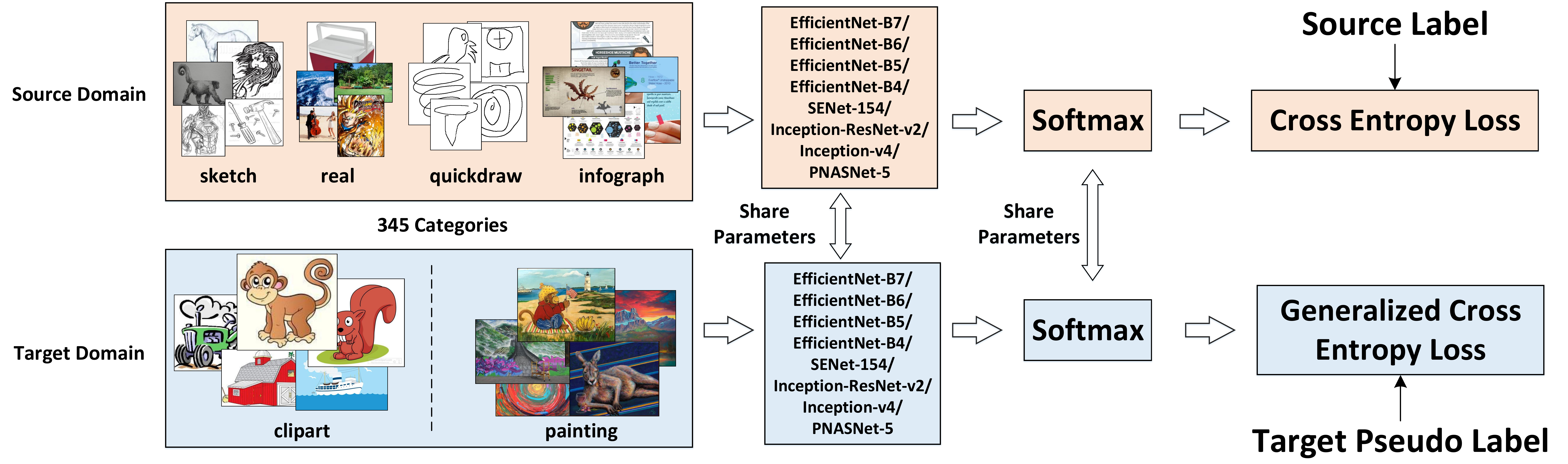}}
\vspace{-0.1in}
	\caption{\small An overview of our End-to-End Adaptation (EEA) module for multi-source domain adaptation task.}
	\label{fig:1}
	\vspace{-0.2in}
\end{figure*}

\begin{figure*}[!tb]
\vspace{-0.0in}
	\centering {\includegraphics[width=0.8\textwidth]{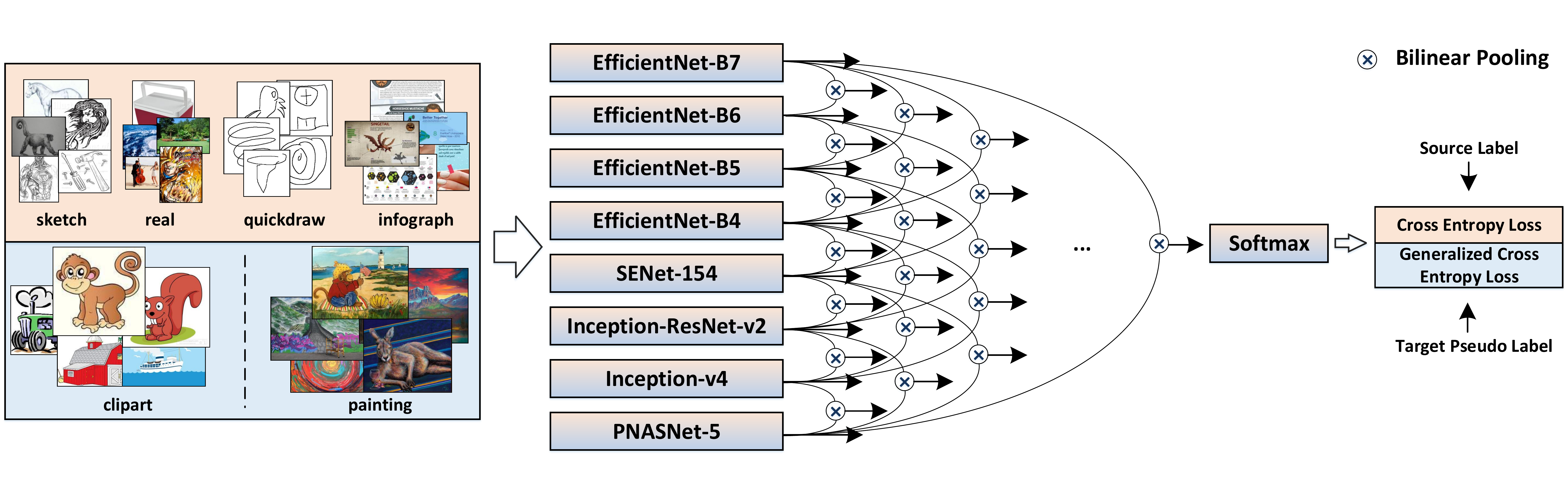}}
\vspace{-0.2in}
	\caption{\small An overview of our Feature Fusion based Adaptation (FFA) module for multi-source domain adaptation task.}
	\label{fig:2}
	\vspace{-0.2in}
\end{figure*}

\section{Multi-Source Domain Adaptation}
Inspired from unsupervised image/video translation \cite{chen2019mocycle,zhu2017unpaired}, we utilize CycleGAN \cite{zhu2017unpaired} to perform unsupervised pixel-level adaptation between source domains (sketch and real) and target domain (clipart/painting), respectively. Thus, each unlabeled training image in sketch or real domains is translated into an image in target domain via the generator of CycleGAN (named as sketch* and real* domains). Figure \ref{fig:0} shows several examples of such pixel-level adaptation from source domains (sketch and real) to target domain (clipart/painting). Next, we combine all the six source domains (sketch, real, quickdraw, infograph, sketch*, and real*) and train eight source-only models in different backbones (EfficientNet-B7 \cite{tan2019efficientnet}, EfficientNet-B6 \cite{tan2019efficientnet}, EfficientNet-B5 \cite{tan2019efficientnet}, EfficientNet-B4 \cite{tan2019efficientnet}, SENet-154 \cite{hu2018squeeze}, Inception-ResNet-v2 \cite{szegedy2017inception}, Inception-v4 \cite{szegedy2017inception}, PNASNet-5 \cite{liu2018progressive}). All backbones are pre-trained on ImageNet and we can achieve the initial pseudo label for each unlabeled target sample by averaging the predictions of eight source-only models. Furthermore, a hybrid system with two kinds of adaptation models (End-to-end adaptation module and Feature fusion based adaptation module) are utilized to fully exploit pseudo labels for this task. We alternate the two adaptation models in four times for enhancing pseudo labels.

\textbf{End-to-End Adaptation Module (EEA).}
This module performs domain adaptation by fine-tuning source-only models with updated pseudo labels in an end-to-end fashion. Figure \ref{fig:1} depicts its detailed architecture. In particular, for unlabeled target data, generalized cross entropy loss \cite{zhang2018generalized} is adopted for training with pseudo labels. After training, we update pseudo labels of unlabeled target samples by averaging the predictions of eight adaptation models in different backbones.

\textbf{Feature Fusion based Adaptation Module (FFA).}
This module directly extracts features from each backbone in the former module and fuses features from every two backbones via Bilinear Pooling. Next, for each kind of fused feature for input source/target sample, we take it as input and train a classifier from scratch. Each classifier is equipped with cross entropy loss (for labeled source sample) and generalized cross entropy loss (for unlabeled target sample). We illustrate this module in Figure \ref{fig:2}. After training the 36 classifiers (28 classifiers with input fused feature and 8 classifiers with input single feature), we update pseudo labels of unlabeled target sample by averaging the predictions of 36 classifiers. At inference, we take the averaged output from 36 classifiers (learnt in Feature fusion based adaptation module at the last time) as the final prediction.

\begin{figure}[!tb]
\vspace{-0.0in}
	\centering {\includegraphics[width=0.5\textwidth]{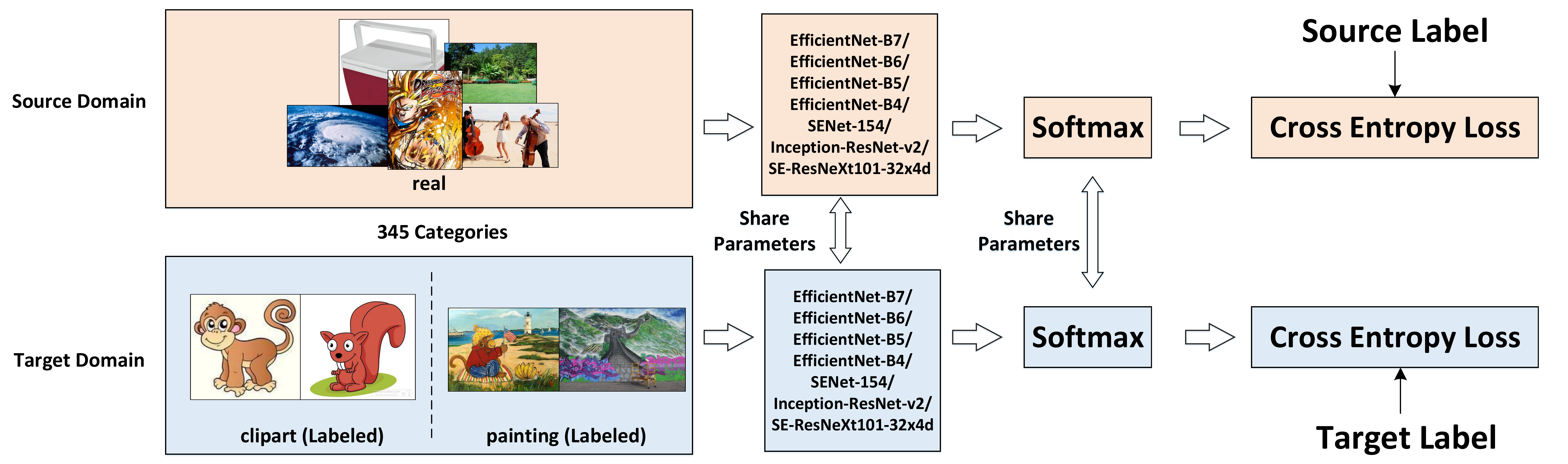}}
	\caption{\small An overview of classifier pre-training for semi-supervised domain adaptation task.}
	\label{fig:3}
	\vspace{-0.10in}
\end{figure}

\begin{figure}[!tb]
\vspace{-0.0in}
	\centering {\includegraphics[width=0.5\textwidth]{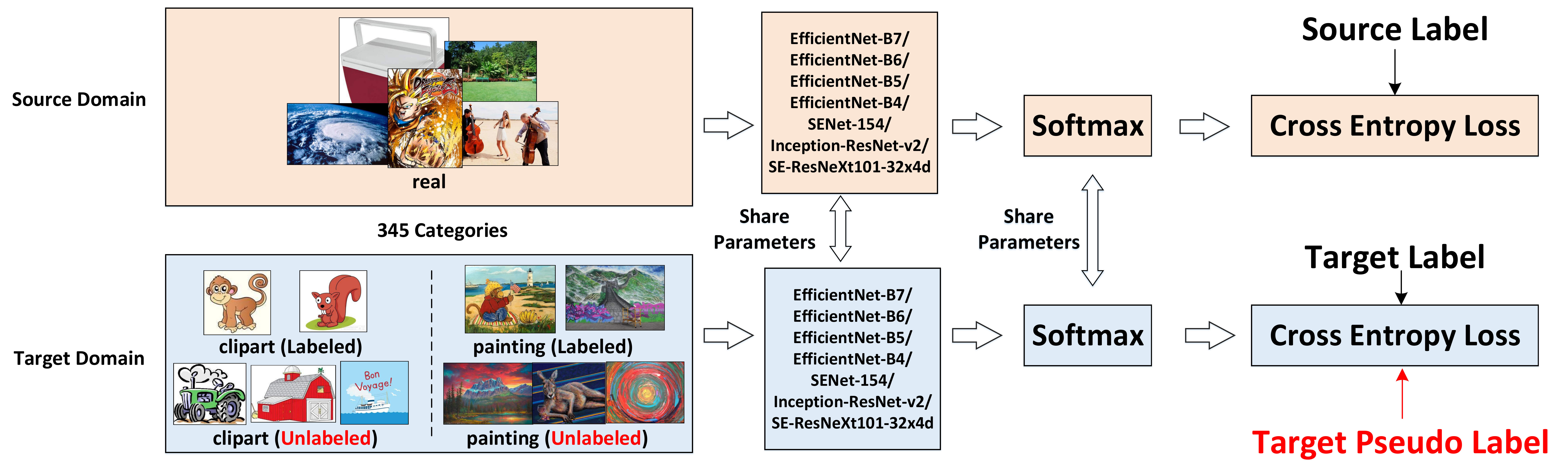}}
	\caption{\small An overview of our End-to-End Adaptation (EEA) module for semi-supervised domain adaptation task.}
	\label{fig:4}
	\vspace{-0.20in}
\end{figure}

\begin{table*}\scriptsize
\centering
\vspace{-0.10in}
\caption{Comparison of different sources and backbones in source-only model for multi-source domain adaptation task on Validation Set.}

\label{table:1}
\begin{tabular}{c|l|c|l|c|c}
\Xhline{0.8pt}
\textbf{Method}~~~&~~~\textbf{~~~~~~~~~~~~~~~Source}~~~&~~~\textbf{Target}~~~&~~~~~~~~~~\textbf{Backbone}~~~&\textbf{mean\_acc\_all}&\textbf{mean\_acc\_classes}\\ \Xhline{0.8pt}
Source-only     & real & sketch & SE-ResNeXt101\_32x4d & 40.24\% &39.59\%\\
Source-only        & real, quickdraw & sketch& SE-ResNeXt101\_32x4d & 43.09\% & 41.76\%\\
Source-only        & real, quickdraw, infograph & sketch& SE-ResNeXt101\_32x4d  & 48.22\% & 46.95\%\\
Source-only        & real, quickdraw, infograph, real* & sketch& SE-ResNeXt101\_32x4d   & \textbf{50.27}\% & \textbf{48.59}\%\\\hline
Source-only        & real, quickdraw, infograph, real* & sketch& Inception-v4  & 51.08\% & 49.22\%\\
Source-only        & real, quickdraw, infograph, real* & sketch& Inception-ResNet-v2 & 52.50\% & 50.94\% \\
Source-only        & real, quickdraw, infograph, real* & sketch& PNASNet-5  & 51.64\% & 49.52\%\\
Source-only        & real, quickdraw, infograph, real* & sketch& SENet-154  & 52.40\% & 50.46\%\\
Source-only        & real, quickdraw, infograph, real* & sketch& EfficientNet-B4  & 53.30\%& 51.82\%\\
Source-only        & real, quickdraw, infograph, real* & sketch& EfficientNet-B6  & 53.85\% & 51.98\%\\
Source-only        & real, quickdraw, infograph, real* & sketch& EfficientNet-B7  & \textbf{54.72}\% & \textbf{52.92}\%\\
\Xhline{0.8pt}
\end{tabular}
\vspace{-0.1in}
\end{table*}

\begin{table*}\scriptsize
\centering
\caption{Comparison of different methods for multi-source domain adaptation task on Validation Set.}
\setlength\tabcolsep{2.5pt}
\label{table:2}
\begin{tabular}{l|l|c|l|c|c}
\Xhline{0.8pt}
\textbf{Method}~~~&~~~\textbf{~~~~~~~~~~~~~~~Source}~~~&~~~\textbf{Target}~~~&~~~~~~~~~\textbf{Backbone}&\textbf{mean\_acc\_all}&\textbf{mean\_acc\_classes}\\ \Xhline{0.8pt}
Source-only     & real, quickdraw, infograph & sketch & ResNet-101 & 43.53\% &42.73\%\\
SWD \cite{lee2019sliced}   & real, quickdraw, infograph & sketch& ResNet-101 & 44.36\% & 43.74\%\\
MCD \cite{saito2018maximum}       & real, quickdraw, infograph & sketch& ResNet-101  & 45.01\% & 44.03\%\\\hline
Source-only        & real, quickdraw, infograph & sketch& SE-ResNeXt101\_32x4d   & 48.22\% & 46.95\%\\
BSP+CDAN \cite{chen2019transferability}       & real, quickdraw, infograph & sketch& SE-ResNeXt101\_32x4d  & 53.01\% & 51.36\%\\
CAN \cite{kang2019contrastive}       & real, quickdraw, infograph & sketch& SE-ResNeXt101\_32x4d & 54.74\% & 52.89\% \\
CAN \cite{kang2019contrastive} +TPN \cite{pan2019transferrable}     & real, quickdraw, infograph & sketch& SE-ResNeXt101\_32x4d  & 56.49\% & 54.43\%\\\hline
End-to-End Adaptation (Cross Entropy)   & real, quickdraw, infograph & sketch& SE-ResNeXt101\_32x4d  & 54.42\% & 53.18\%\\
End-to-End Adaptation (Generalized Cross Entropy)        & real, quickdraw, infograph & sketch& SE-ResNeXt101\_32x4d  & \textbf{58.09}\%& \textbf{56.15}\%\\\Xhline{0.8pt}
\end{tabular}
\vspace{-0.2 in}
\end{table*}

\section{Semi-Supervised Domain Adaptation}
For semi-supervised domain adaptation task, we over-sample the labeled target samples ($\times$10) and combine them with labeled source samples for training classifier in a supervised setting. Figure \ref{fig:3} depicts the detailed architecture for classifier pre-training. Note that here we train seven kinds of classifiers in different backbones (EfficientNet-B7, EfficientNet-B6, EfficientNet-B5, EfficientNet-B4, SENet-154, Inception-ResNet-v2, SE-ResNeXt101-32x4d). All backbones are pre-trained on ImageNet and we can achieve the initial pseudo label for each unlabeled target sample by averaging the predictions of the seven classifiers.

\textbf{End-to-End Adaptation Module (EEA).}
Next, an end-to-end adaptation module is utilized to incorporate pseudo labels for training classifiers (in the backbones pre-trained on ImageNet), which further bridges the domain gap between source and target domain. Figure \ref{fig:4} illustrates this module. After training, we update pseudo labels of unlabeled target samples by averaging the predictions of seven classifiers in different backbones. The updated pseudo labels will be utilized to train the end-to-end adaptation module again. We repeat such procedure for three times.

\textbf{Prototype-based Classification Module (PC).}
Taking the inspiration from Prototype-based adaptation \cite{pan2019transferrable}, we construct an additional non-parametric classifier to strengthen the predictions from the previous EEA module. Specifically, under each backbone, we define the prototype of each class as the average of all labeled target samples in that class (according to the given labels and pseudo labels). Therefore, the prototype-based classification for each target sample is performed by measuring the distances to prototypes of each class.
At inference stage, we take the averaged output from 1) seven classifiers learnt in end-to-end adaptation module at the last time and 2) seven prototype-based classifiers as the final prediction.

\begin{table}\scriptsize
\centering
\setlength\tabcolsep{2.5pt}
\caption{Comparison of different backbones in our End-to-End Adaptation (EEA) module and Feature Fusion based Adaptation (FFA) module for multi-source domain adaptation on Validation Set (Source: real, quickdraw, infograph, real*; Target: sketch).}
\label{table:3}
\begin{tabular}{l|l|c|c}
\Xhline{0.8pt}
\textbf{Method}&~~~~~~~~~\textbf{Backbone}&\textbf{mean\_acc\_all}&\textbf{mean\_acc\_classes}\\ \Xhline{0.8pt}
EEA &SE-ResNeXt101\_32x4d&	59.07\%&	57.05\%\\
EEA &Inception-v4	&59.93\%&	57.42\%\\
EEA &Inception-ResNet-v2	&60.58\%	&58.32\%\\
EEA &PNASNet-5	&60.07\%	&57.84\%\\
EEA &SENet-154	&60.88\%	&58.29\%\\
EEA &EfficientNet-B4	&60.41\%	&58.30\%\\
EEA &EfficientNet-B6	&61.12\%	&58.80\%\\
EEA &EfficientNet-B7	&63.01\%	&60.33\%\\\hline
FFA& Ensemble	&\textbf{67.58}\%	&\textbf{64.54}\%\\
\Xhline{0.8pt}
\end{tabular}
\vspace{-0.2in}
\end{table}

\begin{table*}\small
\centering
\vspace{-0.15in}
\setlength\tabcolsep{3.5pt}
\caption{Comparison of different components in our system for multi-source domain adaptation on Testing Set (Source: sketch, real, quickdraw, infograph, sketch*, real*; Target: clipart/painting).}
\label{table:4}
\begin{tabular}{l|l|c|c|c}
\Xhline{0.8pt}
\textbf{Method}&\textbf{~~~~~~Backbone}&\textbf{mean\_acc\_all (clipart)}&\textbf{mean\_acc\_all (painting)}&\textbf{mean\_acc\_all}\\ \Xhline{0.8pt}
Source-only & Inception-ResNet-v2 & 67.77\%&	59.24\%	&62.59\%\\\hline
EEA+FFA & Ensemble & 78.16\%	&67.56\%	&71.73\%\\
(EEA+FFA)$^2$& Ensemble &79.66\%	&69.51\%	&73.50\%\\
(EEA+FFA)$^2$, Higher resolution & Ensemble &81.25\%	&71.65\%	&75.42\%\\
(EEA+FFA)$^4$, Higher resolution & Ensemble &\textbf{81.61}\%	&\textbf{72.31}\%	&\textbf{75.96}\%\\
\Xhline{0.8pt}
\end{tabular}
\vspace{-0.2in}
\end{table*}

\begin{table}\small
\centering
\caption{Comparison of different components in our system for semi-supervised domain adaptation on Testing Set (Source: real; Target: clipart/painting).}
\label{table:5}
\begin{tabular}{l|l|c}
\Xhline{0.8pt}
\textbf{Method}&\textbf{Backbone}&\textbf{mean\_acc\_all}\\ \Xhline{0.8pt}
Source-only &Ensemble &64.3\%\\\hline
EEA&Ensemble &68.8\%\\
EEA$^2$&Ensemble &70.5\%\\
EEA$^3$&Ensemble &71.35\%\\
EEA$^3$+PC &Ensemble &\textbf{71.41}\%\\
\Xhline{0.8pt}
\end{tabular}
\vspace{-0.25in}
\end{table}

\section{Experiments}
\subsection{Multi-Source Domain Adaptation}

\textbf{Effect of pixel-level adaptation in source-only model.} Compared to traditional UDA, the key difference in multi-source domain adaptation task is the existence of multiple sources. To fully explore
the effect of multiple source domains and the synthetic domain via pixel-level adaptation, we show the performances of source-only model on validation set by injecting one more source domain in Table \ref{table:1}. The results across different metrics consistently indicate the advantage of transferring knowledge from multiple source domains. The performance is further improved by incorporating synthetic domain (real*) via pixel-level adaptation. Table \ref{table:1} additionally shows the performances of source-only model under different backbones and the best performance is observed when we construct source-only model under EfficientNet-B7.

\textbf{Effect of End-to-End Adaptation (EEA).} We evaluate our End-to-End Adaptation module on Validation Set and compare the results to recent state-of-the-art UDA techniques (e.g., SWD \cite{lee2019sliced}, MCD \cite{saito2018maximum}, BSP+CDAN \cite{chen2019transferability}, CAN \cite{kang2019contrastive}, and TPN \cite{pan2019transferrable}). Results are presented in Table \ref{table:2}. Overall, our adopted EEA with Generalized Cross Entropy exhibits better performance than other runs, which demonstrates the merit of self-learning for multi-source domain adaptation. Note that here we include one variant of our EEA by replacing Generalized Cross Entropy with traditional Cross Entropy, which results in inferior performance. The results verify the advantage of optimizing classifier with Generalized Cross Entropy for unlabeled target samples in self-learning paradigm.

\textbf{Effect of Feature Fusion based Adaptation (FFA).} One of the important design in our system is feature fusion based adaptation (FFA) which facilitate the learning of domain-invariant classifier with fused features from different backbones. As shown in Table \ref{table:3}, by fusing the features from every two backbones in EEA via Bilinear Pooling, our FFA leads to a large performance improvement.

\textbf{Performance on Testing Set.} Table \ref{table:4} illustrates the final performances of our submitted systems with different settings on Testing Set. The basic component in our submitted systems is the hybrid system consisting of two adaptation modules (EEA and FFA), which will be alternated in several times. For simplicity, we denote the system which alternates (EEA+FFA) in $N$ times as (EEA+FFA)$^N$. Note that we also try to enlarge the input resolution of each backbone (+ 64 pixels in both width and hight) in the submitted systems and such processing is named as ``Higher resolution." As shown in Table \ref{table:4}, our system with more alternation times and Higher resolution achieves the best performance on Testing Set.

\subsection{Semi-Supervised Domain Adaptation}
The performance comparisons between our submitted systems for semi-supervised domain adaptation task on Testing Set are summarized in Table \ref{table:5}. Note that here we denote the setting which alternates End-to-End Adaptation (EEA) module in $N$ times as EEA$^N$. In general, our system with more alternation times obtains higher performance. In addition, by fusing the predictions from both EEA and Prototype-based Classification (PC), our system boosts up the performance.

{\small
\bibliographystyle{ieee_fullname}
\bibliography{egbib}
}

\end{document}